\documentclass{article}

\usepackage[final]{neurips_bdl2019}

\usepackage[utf8]{inputenc} 
\usepackage[T1]{fontenc}    
\usepackage{hyperref}       
\usepackage{url}            
\usepackage{booktabs}       
\usepackage{amsfonts}       
\usepackage{nicefrac}       
\usepackage{microtype}      
\usepackage{amsmath, amsfonts, amssymb}
\usepackage{color}
\usepackage{subfigure}
\usepackage{graphicx}

\usepackage{xfrac}
\usepackage{mathtools}

\title{Stabilising priors for robust Bayesian deep learning}

\author{
  Felix McGregor\thanks{Correspondence: flx.mcgregor@gmail.com}  \\
  Department of E\&E Engineering  \\
  Stellenbosch University
  \And
  Arnu Pretorius \\
  Computer Science Division \\
  Stellenbosch University
  \And
  Johan du Preez \\
  Department of E\&E Engineering \\
  Stellenbosch University\\
  \And
  Steve Kroon \\
  Computer Science Division \\
  Stellenbosch University\\
}

\begin{document}

\maketitle

\begin{abstract}
    Bayesian neural networks (BNNs) have developed into useful tools for probabilistic modelling due to recent advances in variational inference enabling large scale BNNs. However, BNNs remain brittle and hard to train, especially: (1) when using deep architectures consisting of many hidden layers and (2) in situations with large weight variances. We use signal propagation theory to quantify these challenges and propose \textit{self-stabilising priors}. 
    This is achieved by a reformulation of the ELBO to allow the prior to influence network signal propagation. Then, we develop a stabilising prior, where the distributional parameters of the prior are adjusted before each forward pass to ensure stability of the propagating signal. This stabilised signal propagation leads to improved convergence and robustness making it possible to train deeper networks and in more noisy settings. 
\end{abstract}
\section{Introduction}
%
%

Bayesian neural networks (BNNs) offer a way of combining uncertainty estimation with the expressive power of deep learning. 
Advances in BNNs \citep{graves2011practical, blundell2015weight, kingma2013auto, rezende2014stochastic} have made it possible to scale these models but not yet to the level of success of modern deep learning.
The reason in part, is that BNNs are difficult to train. 
Stochastic optimisation methods for these models tend to exhibit high variance and BNNs can be very sensitive to small changes in hyperparameters, architecture and the choice of prior.
This work builds on \citet{pretorius2018critical}, which extended the analysis of signal propagation for deterministic ReLU networks to ReLU networks with stochastic regularisation. 
In light of recent links between stochastic regularisation techniques and variational inference \citep{gal2016dropout, lrt}, we relate the initialisation techniques derived in \cite{pretorius2018critical}, to an iteratively updating prior to stabilise the flow of information through ReLU BNNs throughout training.
Signal propagation analysis of BNNs in the infinite width limit leads us to propose \textit{self-stabilising priors} for robust training. 
We design an iterative prior with distributional parameters derived to preserve the variance of signals propagating forward through the network. 
This is a heuristic similar in nature to an iterated application of Empirical Bayes~(EB) for setting prior hyperparameters, as is common in BNN training (\cite{graves2011practical} and \cite{titsias2014doubly} propose closed form updates for prior hyperparameters). While EB chooses hyperparameters that optimize the likelihood, our approach chooses prior hyperparameters for each forward pass that attempt to optimize signal propagation behaviour in the network.
In order for this prior to influence the signal propagation dynamics, we reformulate the \textit{evidence lower bound} (ELBO) objective to allow this prior to exercise its stabilising effect during the forward pass. 



\section{Self-stabilising priors}

The prior $p_\alpha(W)$ usually impacts the ELBO through an additive KL term, only affecting the weights after the forward pass, having no effect on the signal propagation dynamics of the network. 
We instead propose priors that also exert their influence \textit{during} the forward pass, so as to promote stable signal propagation, and improve robustness in deep BNNs.
We reformulate the ELBO in terms of a combination of the prior and approximating posterior as $\tilde{q}_{\{\alpha, \phi\}}(W) = p_\alpha(W)q_\phi(W) / Z$. 
This formulation allows us to estimate an expectation using a Monte Carlo estimator with samples drawn from $\tilde{q}_{\{\alpha, \phi\}}(W)$ instead of $q_\phi(W)$. This ensures that the sampled weights of the network are being influenced by the current prior $p_\alpha(W)$ during the forward pass. From Appendix B we define the ELBO as
\begin{align}
    \label{eq: adelbo reparam}
    \mathcal{L}_{\tilde{q}} \coloneqq \mathbb{E}_{p(\epsilon)} \left [  \log p(\mathbf{y}|\mathbf{x}, \mathbf{b}, W = \xi(\epsilon, \alpha, \phi)) \right] -  \textrm{KL}(\tilde{q}_{\{\alpha, \phi\}}(W) || p_\alpha(W)),
\end{align}
where $\epsilon \sim \mathcal{N}(0,I)$. The prior is adjusted after every gradient update to adapt to the updated posterior $q_\phi(W)$. 
These prior parameters are optimal in the sense that together with the reformulated ELBO in \eqref{eq: adelbo reparam}, the sampled weights from $\tilde{q}_{\{\alpha, \phi\}}(W)$ have a stabilising effect on the signal propagation dynamics of a Bayesian deep neural network. 

To quantify the signal propagation dynamics we make use of the \textit{mean field} assumption \citep{saxe2013exact, poole2016exponential} which allows for the components of the pre-activation vectors $\mathbf{h}^l$ to be treated as independent Gaussian random variables. 
Then, according to the central limit theorem, these pre-activations are Gaussian distributed. We consider independent Gaussian priors $p_\alpha(w_{i,j}^l) = \mathcal{N}(\mu_{p_{ij}}^l, (\sigma_{p_{i,j}}^l)^2)$ and posteriors $q_\phi(w_{i,j}) = \mathcal{N}(\mu_{q_{ij}}^l, (\sigma_{q_{i,j}}^l)^2)$. Focusing on ReLU activations, i.e. $g(a) = \max(0, a)$, we consider a single scalar hidden unit pre-activation $h^l_j$ at an arbitrary layer $l$ of the network. In Appendix A, we show that in expectation over the weights and biases,
under the assumption of zero mean inputs at each layer (true at initialisation but a somewhat unrealistic assumption during training further discussed in Appendix A),
the variance of $h^l_j$, governing signal propagation in the forward pass is given by
\begin{align}
\textrm{Var}[h^l_j] & = \left [ \left (1-\frac{1}{\pi} \right)(\mu^l_{\tilde{q}_j})^2 + (\sigma^l_{\tilde{q}_j})^2 \right ]\frac{\textrm{Var}[h^{l-1}_{j^{\prime}}]}{2}. \label{eq: pre-act variancea}
\end{align}
where $j^{\prime} \in \{1,...,D_{l-1}\}$. Then, to stabilise the signal, we want to find prior parameters $\alpha=\{\mu_{p_{ij}}^l, \sigma_{p_{i,j}}^l\}$ that can preserve the variance during the forward pass. Specifically, we want $\alpha$ to ensure $\textrm{Var}[h^l_j] = \textrm{Var}[h^{l-1}_{j^{\prime}}]$ in \eqref{eq: pre-act variancea}. 
To achieve this, we require $\mu^l_{p_{ij}} = \mu^l_{q_{ij}}$ 
giving the stabilising prior parameters $\alpha$ as
\begin{align}
\label{eq: optimal prior parameters}
\mu_{p_{ij}}^l = \mu_{q_{ij}}^l, \textcolor{white}{white space} \sigma_{p_j}^l & = \sqrt{\frac{ (\sigma^l_{\tilde{q}_j})^2 \gamma}{(\sigma^l_{\tilde{q}_j})^2 - \gamma}} ,
\end{align}
where $\gamma = |2 - (1 - 1/\pi)(\mu^l_{q_j})^2|$ and we take the absolute value to ensure positive variances. 
Note that we can apply the result in \eqref{eq: pre-act variancea} recursively for all layers $l = 1, ..., L$, with base case $\textrm{Var}[x^0] = \frac{1}{D_0}\mathbf{x}^0 \cdot \mathbf{x}^0$. 
Therefore, sampling the weights $W \sim \tilde{q}_{\{\alpha, \phi\}}(W)$ at each forward pass, while setting the prior $p_\alpha(w) = \mathcal{N}(\mu_q^l, | (\sigma^l_q)^2 \gamma/ ((\sigma^l_q)^2 - \gamma)| / D_{l-1})$, promoting stable signal propagation. 

The effect of the stabilising prior is that the larger the mean and variance of the incoming weights, the more likely it is to destroy the signal, whereas if the second moment is small, it is not likely to add noise to the signal. The prior encourages the weight distribution to sample closer to its means when the second moment of the distribution is large.



\section{Experiments}

\begin{figure*}[!htp]
	\centering
	
	\centering
	\includegraphics[width=0.8\textwidth, angle=0]{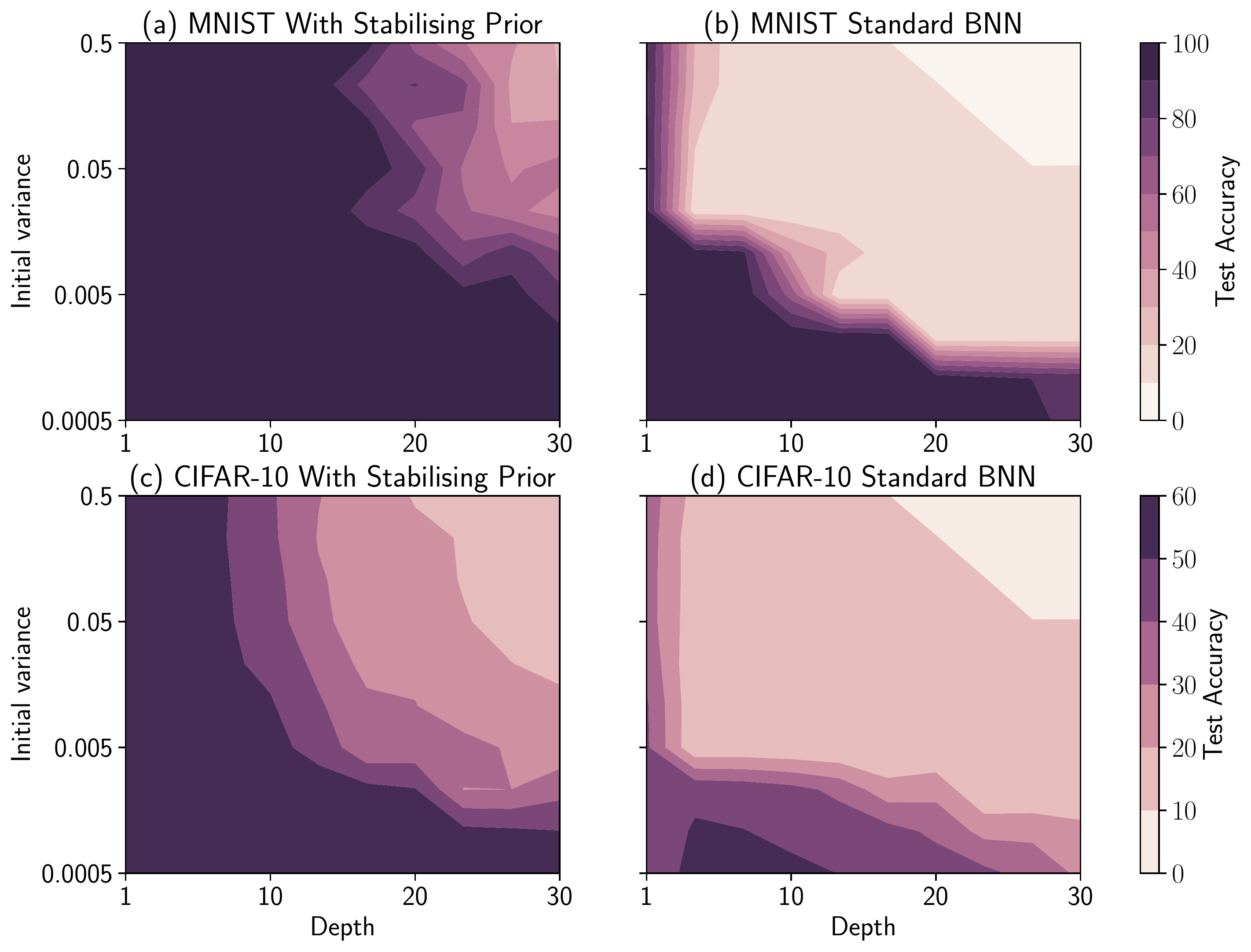} 
	
	\caption{MNIST and CIFAR-10 large scale experiments. Classification accuracy grid of ReLU networks trained with varying depths and initial variance conditions trained for 50 epochs.}
	\label{robust}
\end{figure*}

\textbf{Limits of trainability.}\hspace{1mm} We restrict ourselves to BNNs with fully connected layers with a specified number of hidden layers all of constant width. \footnote{Code available \url{https://git.io/JeLkH}}. We investigate performance at extremities by training a series of networks with varying depths and initial variances. We compare a series of networks with our proposed stabilising prior incorporated on the forward pass with a standard non-conjugate Gaussian prior \cite{titsias2014doubly} which we report in Figures \ref{robust} (a) and (b). We observe our stabilising prior makes it possible to train deeper BNNs and in more noisy conditions.

\begin{figure}[!htp]
	\centering
	\includegraphics[width=0.8\textwidth, angle=0]{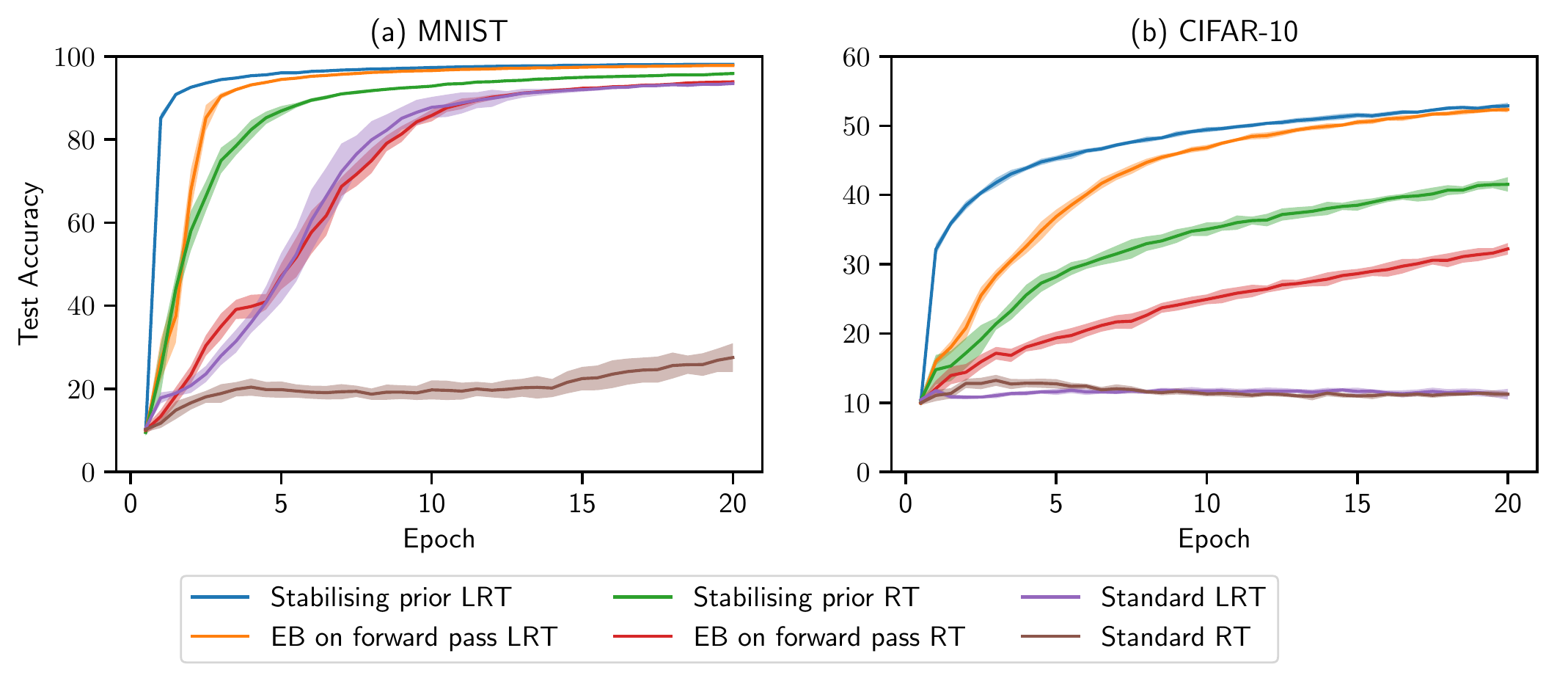} 
	\caption{Progression of test accuracy for various networks through training averaged over 10 runs for a 5 layer deep 512 wide network with an initial variance of 0.001.}
	\label{money}
\end{figure}

\textbf{Accelerated training.} \hspace{1mm} In general, we also observe that our prior improves convergence as demonstrated in Figure \ref{money}. We compare with Empirical Bayes (EB) as in \cite{titsias2014doubly} which uses the gradient to find optimal hyperparameters for the prior. 
We further compare these priors with a Gaussian prior and report their results for both the reparametrisation trick (RT) \cite{kingma2013auto} and local reprametrisation trick (LRT) \cite{lrt}. 

\section{Discussion}

We used signal propagation theory to derive priors for BNNs that promote stable signal propagation. The prior incorporates knowledge of model architecture and activation function derived from how signals propagate in the network in the infinite width limit. 
We showed that these priors, when their effect is exerted in the forward pass, makes it possible to train deeper networks and in more noisy conditions. 
This alleviates the need to tune hyper-parameters and extends BNNs to deeper architectures. 
We also observe that stable signal propagation accelerates training, which we attribute to cleaner signals and gradients being propagated through the network and more efficient expectations.



\section*{Acknowledgements}
We would like to thank the Technology and Human Resource for Industry Programme (THRIP) as well as Centre for Artificial Intelligence
Research (CAIR) for financial support.
We gratefully acknowledge the support of NVIDIA Corporation with the donation of the Titan Xp GPU used for this research.


\bibliographystyle{IEEEtranN}
\bibliography{bibfile}

\appendix

\section*{Appendices}
\section{Signal propagation in BNNs}

We consider the quantity $\mathbb{E}[(h^l_j)^2]$ during a forward pass. This quantity is the second moment of a single hidden unit $h^l_j$ in layer $l$, consisting of $D_{l-1}$ incoming connections, where the expectation taken is over the network parameters and is given by 
\begin{align}
    \label{eq: stable prior 1}
    \mathbb{E}[(h^l_j)^2] & = \mathbb{E}\left[\left(\sum^{D_{l-1}}_{i=1}w_{ij}^lx_i^l + b^l_j\right)^2\right] \nonumber \\
    & = \sum^{D_{l-1}}_{i=1}\mathbb{E}[(w^l_{ij})^2]\mathbb{E}[(x^{l}_i)^2] + \mathbb{E}[(b^l_j)^2].
\end{align}
We define
\begin{align}
\sigma_{\tilde{q}_j}^l = \sum_{i=1}^{D_{l-1}} (\sigma^l_{{ij}})^2 \qquad\text{and}\qquad & \mu_{\tilde{q}_j}^l = \sum_{i=1}^{D_{l-1}}\mu_{{ij}}^l
\end{align}
to use as the statistics to describe $\sum_{i=1}^{D_{l-1}}w_{ij}$. Note that while it is true that we can write 
\begin{align}
	\sum_{i=1}^{D_{l-1}}w^l_{ij} \sim \mathcal{N}\left(\mu_{\tilde{q}_j}^l, (\sigma_{\tilde{q}_j}^l)^2\right)
\end{align}
at initialisation, in our analysis of the network at an arbitrary stage of the stage of training the i.i.d. assumption of the central limit theorem (CLT) does not strictly hold. As in \cite{Wu}, we empirically find that some form of the CLT holds for the hidden units during training. We thus continue to approximate the expectation with a Gaussian according to the CLT. 

Next, in designing $\tilde{q}_{\beta}(W)$ we scale the variance and impose $\mathbb{E}[(w^l_{ij})^2] = \frac{(\mu_{\tilde{q}}^l)^2 + (\sigma_{\tilde{q}}^l)^2}{D_{l-1}}, \forall i,j$ to ensure that the variance is bounded in the infinite width limit \cite{schoenholz2016deep}. This also allows the variance propagated forward to be independent of the layer width. We now have 
\begin{align}
    \label{eq: stable prior 2}
    \mathbb{E}[(h^l_j)^2] & = ((\mu_{\tilde{q}_j}^l)^2 + (\sigma_{\tilde{q}_j}^l)^2)\frac{1}{D_{l-1}}\sum^{D_{l-1}}_{i=1}g(h^{l-1}_i)^2 +((\mu_b^l)^2+ (\sigma_b^l)^2).
\end{align}
As $D_{l-1} \rightarrow \infty$, $h^l_j$ becomes an infinite sum of i.i.d.\ random variables and becomes Gaussian distributed according to the CLT. We can thus write
\begin{align}
    \label{eq: stable prior 3}
    \mathbb{E}[(h^l_j)^2] & = ((\mu_{\tilde{q}}^l)^2 + (\sigma_{\tilde{q}_j}^l)^2)\mathbb{E}_z[\phi(\tau^{l-1} + \sqrt{\nu^{l-1}}z)^2] + (\mu_b^l)^2+ (\sigma_b^l)^2
\end{align}
where $z \sim \mathcal{N}(0, 1)$, and $\tau^{l-1}$ and $\nu^{l-1}$ are the incoming signal to layer $l$'s mean and variance respectively. 
If we use ReLU as activation, i.e. $g(a) = \max(0, a)$, then
\begin{align}
    \label{eq: stable prior 4}
    \mathbb{E}[(h^l_j)^2] & = ((\mu_{\tilde{q}}^l)^2 + (\sigma_{\tilde{q}_j}^l)^2)\left \{ \int_{-\infty}^\infty \Phi(z)\phi(\tau^{l-1} + \sqrt{\nu^{l-1}}z)^2 dz \right \} + (\mu_b^l)^2+ (\sigma_b^l)^2 \nonumber\\
    & = ((\mu_{\tilde{q}}^l)^2 + (\sigma_{\tilde{q}_j}^l)^2) 
     \left \{ \int_{0}^\infty \Phi(z) \left((\tau^{l-1})^2 + 2\tau^{l-1}\sqrt{\nu^{l-1}}z+ \nu^{l-1}z^2\right)  dz \right \}\nonumber \\
     & \qquad \qquad\qquad\qquad\qquad\qquad\qquad\qquad\qquad\qquad\qquad\qquad\qquad+(\mu_b^l)^2+ (\sigma_b^l)^2 \nonumber \\
    & = ((\mu_{\tilde{q}}^l)^2 + (\sigma_{\tilde{q}_j}^l)^2) \left[\frac{(\tau^{l-1})^2}{2} + \frac{2\tau^{l-1}\sqrt{\nu^{l-1}}}{\sqrt{2 \pi}} + \frac{\nu^{l-1}}{2}\right] +(\mu_b^l)^2+ (\sigma_b^l)^2
\end{align}
where $\Phi(z) = \frac{e^{-z^2/2}}{\sqrt{2\pi}}$. 

Similarly, we can show that the relevant statistics governing signal propagation in the forward pass are given by
\begin{align}
\mathbb{E}[h^l_j] & = \mathbb{E}\left[\sum^{D_{l-1}}_{j=1}w^l_{ij}g(h^{l-1}_j + b_i)\right] \nonumber \\
& = \mu_{\tilde{q}_j}^l\bigg(\frac{\tau^{l-1}}{2}+\sqrt{\frac{\nu^{l-1}}{2\pi}} \bigg) + \mu_b^l \label{eq: pre-act mean}
\end{align} 
and
\begin{align}
\nu^l_j & = \mathbb{E}\left[\left(\sum^{D_{l-1}}_{j=1}w^l_{ij}g(h^{l-1}_j)\right)^2\right] -  \mathbb{E}\left[\sum^{D_{l-1}}_{j=1}w^l_{ij}g(h^{l-1}_j)\right]^2 \nonumber \\
& = (\mu^l_{\tilde{q}_j})^2 \left[ \frac{(\tau^{l-1})^2}{4}+\tau^{l-1} \sqrt{\frac{\nu^{l-1}}{2\pi}}  + \left(1-\frac{1}{\pi}\right) \frac{\nu^{l-1}}{2}\right]  \nonumber\\
&\qquad + (\sigma^l_{\tilde{q}_j})^2 \left[ \frac{(\tau^{l-1})^2}{2} + 2 \tau^{l-1} \sqrt{\frac{\nu^{l-1}}{2\pi}} + \frac{\nu^{l-1}}{2} \right ] + (\sigma_b^l)^2. \label{eq: pre-act variance}
\end{align} 

Described above is the signal propagation dynamics in general. With a mean preserving prior we can only control variance by multiplicatively expanding or squeezing $\sigma^l_{\tilde{q}_j}$. We can only design for conditions that set $\tau^{l-1}=0$ and  $\sigma_b = 0$ which is true at initialisation but starts to break down during training. In order to continue we thus implicitly assume that during training: (1) the mean of the summed weights' means across a hidden layer's pre-activation remain mean zero i.e. $\mathbb{E}[(\sum_{j=1}^{D_{l}} \mu^l_{q_j})]=0$; (2) biases are zero (note, it is possible to absorb the biases by augmenting the input at each layer with an additional column of ones, this yields more stable signal propagation. We find that treating biases as deterministic parameters aids in training and outweighs the minor gain in stabilising propagation). This allows us to write the variance $\nu^l_j$ as
\begin{align}
	\nu^l_j = \left [ \left (1-\frac{1}{\pi} \right)(\mu^l_{\tilde{q}_j})^2 + (\sigma^l_{\tilde{q}_j})^2 \right ]\frac{\nu^{l-1}}{2}.
	\label{simplesignal}
\end{align}

From here we can design $\tilde{q}_{\{\alpha, \phi\}}(W)$ to preserve variances through the network. We find actually forcing our assumptions and setting parameter means and biases to zero does not train.

The parameters for the joint distribution $\tilde{q}_{\{\alpha, \phi\}}(W)$ are determined by $\alpha$ and $\phi$ as the product of two Gaussian pdfs as
\begin{align}
\label{eq: Gaussian product parameters}
\mu_{\tilde{q}_{ij}}^l = \frac{\mu^l_{q_{ij}}(\sigma^l_{p_{ij}})^2+ \mu^l_{p_{ij}}(\sigma^l_{q_{ij}})^2}{(\sigma^l_{p_j})^2 + (\sigma^l_{q_{ij}})^2}, \textcolor{white}{white space}
\sigma^l_{\tilde{q}_{ij}} = \sqrt{\frac{(\sigma^l_{p_j})^2(\sigma^l_{q_{ij}})^2}{(\sigma^l_{p_j})^2 + (\sigma^l_{q_{ij}})^2}}.
\end{align}

To stabilise the signal, we want to find prior parameters $\alpha \in \{\mu^l_{p_{ij}}, (\sigma^l_{p_j})^2\}$ that can preserve the variance during the forward pass. Specifically, we want $\alpha$ to ensure $\nu^l_j = \nu^{l-1}_j$ or $\textrm{Var}[h^l_j] = \textrm{Var}[h^{l-1}_{j^{\prime}}]$ in \eqref{eq: pre-act variancea} where $j^{\prime} \in \{1,...,D_{l-1}\}$. 
To achieve this, we require $\mu^l_{p_{ij}} = \mu^l_{q_{ij}}$.
Secondly, we find variance parameters of $(\sigma^l_{p_j})^2$ using \eqref{eq: Gaussian product parameters},
to satisfy the condition $(1-1/\pi)(\mu^l_{\tilde{q}_j})^2 + (\sigma^l_{\tilde{q}_j})^2 = 2$ found by setting $\nu^l_j = \nu^{l-1}_j$ in \ref{simplesignal}.


\section{Reformulating the ELBO}

We reformulate the ELBO by lower bounding the log marginal likelihood of the data as follows   
\begin{align}
    \label{eq: elbo 2}
    \log p(\mathbf{y} | \mathbf{x}) & = \log \int p(\mathbf{y}|\mathbf{x}, W)p_\alpha(W)dW \nonumber \\
    & = \log \int p(\mathbf{y}|\mathbf{x}, W)p_\alpha(W) \frac{q_\phi(W)}{q_\phi(W)}dW,
\end{align}
where we combine the prior and approximating posterior as $\tilde{q}_{\{\alpha, \phi\}}(W) = p_\alpha(W)q_\phi(W)/Z$, where $Z$ is a normalisation constant. Then, we can construct a lower bound making use of Jensen's inequality which gives
\begin{align}
    \label{eq: elbo 3}
    \log p(\mathbf{y} | \mathbf{x}) & = \log \int \tilde{q}_{\{\alpha, \phi\}}(W) \frac{p(\mathbf{y}|\mathbf{x}, W)} {q_\phi(W)} ZdW \nonumber \\
    & \geq \int \tilde{q}_{\{\alpha, \phi\}}(W)  \log \frac{p(\mathbf{y}|\mathbf{x}, W)} {q_\phi(W)} ZdW \nonumber\\
    & =  \mathbb{E}_{\tilde{q}_{\{\alpha, \phi\}}(W)} \left [ \log p(\mathbf{y}|\mathbf{x}, W) \right ] - \mathbb{E}_{\tilde{q}_{\{\alpha, \phi\}}(W)} \left [ \log q_\phi(W) \right ] + \log Z. 
\end{align}
By reformulating the ELBO in this way, we can estimate the above expectations using a Monte Carlo estimator with samples drawn from $\tilde{q}_{\{\alpha, \phi\}}(W)$ instead of $q_\phi(W)$. This ensures that the sampled weights of the network are being influenced by the current prior $p_\alpha(W)$ during the forward pass. Finally, we ignore the constant term in \eqref{eq: elbo 3} and replace the cross-entropy term, which is the second term in the equation, with a KL term by introducing the identity ratio $p_\alpha(W)/p_\alpha(W)$ in \eqref{eq: elbo 2}. This gives the following objective when reparametrised using the reparametrisation trick:
\begin{align}
\label{eq: adelbo reparama}
\mathcal{L}_{\tilde{q}} \coloneqq \mathbb{E}_{p(\epsilon)} \left [  \log p(\mathbf{y}|\mathbf{x}, \mathbf{b}, W = \xi(\epsilon, \alpha, \phi)) \right] -  \textrm{KL}(\tilde{q}_{\{\alpha, \phi\}}(W) || p_\alpha(W)),
\end{align}
where $\epsilon \sim \mathcal{N}(0,I)$. 
The prior $p_\alpha(W)$ usually impacts the ELBO through an additive KL term, only affecting the weights after the forward pass, having no effect on the signal propagation dynamics of the network. 
We instead propose priors that also exert their influence \textit{during} the forward pass, so as to promote stable signal propagation, and improve robustness in deep BNNs.

\section{Use of $\tilde{q}(W)$ at test time}
We opt to always include the prior during test time: we sample from $\tilde{q}(W)$ instead of $q(W)$, defining a new posterior of interest, because we find that the accuracy of the model progresses faster.
Since we require adjustment during the forward pass throughout training to ensure stable signal propagation (it is often necessary for any training to occur), it seems reasonable to include it in the forward pass at test time. 
The use of $q(W)$ exhibits performance similar to networks with unstable signal propagation.
Note that the adjustment to the variational posterior for signal propagation is unlike the use of EB to choose the hyperparameters maximizing the likelihood, in that case the prior should not be factored in and $q(W)$ would be appropriate. We investigate what effect including the prior has on the quality of prediction uncertainty in the experiments in Appendix D.

\section{Appendix Experiments: Signal Propagation and Quality of Uncertainty}
\begin{figure*}[!htp]
	\centering
	\includegraphics[width=0.7\textwidth, angle=0]{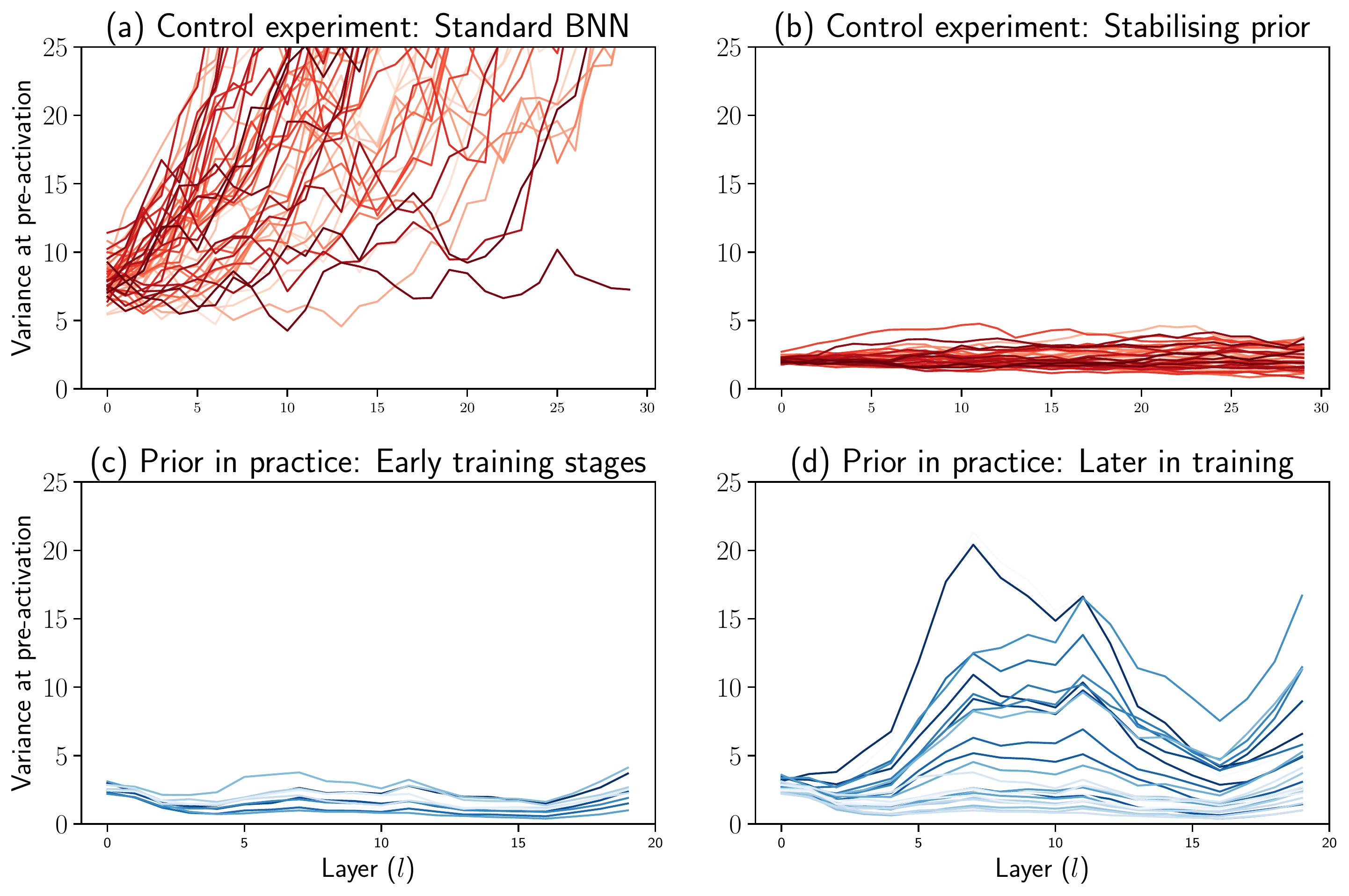} 
	\caption{Signal propagation dynamics of the same signal propagated through different networks. We track the variance of the same data point throughout training by calculating the empirical variance of the vector at the pre-activation at each layer. 
		In a controlled setting we can achieve perfectly stable signal propagation. In practice our assumptions hold for the early stages of training.}
	\label{signalprop}
\end{figure*}

\textbf{Signal propagation.}\hspace{1mm}We examine signals in a ReLU network in Figure \ref{signalprop} to analyse the effect of the prior on the network signal propagation. We monitor the variance dynamics of the same data point throughout training by calculating the empirical variance of the vector of pre-activations at each layer during a forward pass. In Figures \ref{signalprop} (a) and (b) we show a controlled example where we force our assumptions, setting biases and parameter means to zero and see that the prior preserves the signal, whereas in a standard BNN it explodes. Furthermore, we show a typical training scenario in Figures \ref{signalprop} (c) and (d), where we see that our assumptions hold in the early stages of training and start to break down later in training, yielding less stable signal propagation.

\begin{figure*}[!hp]
	\centering
	\includegraphics[width=0.68\textwidth, angle=0]{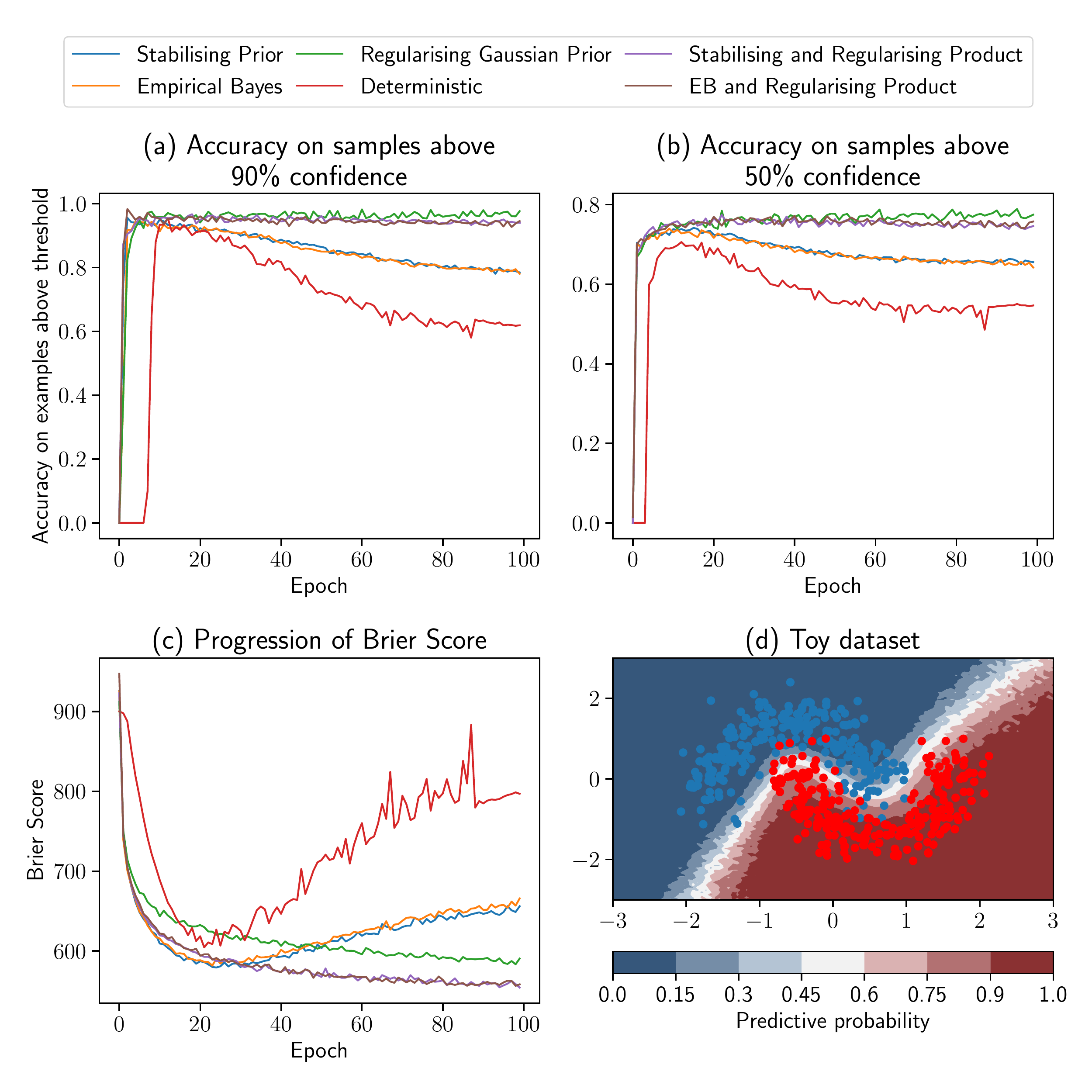} 
	
	\caption{Uncertainty and calibration experiments on CIFAR-10. Iteratively updating priors overfit, however, we can combine regularising and optimal priors to maintain calibrated confidence and better Brier scores. In (d) we also see we are able to get reasonable uncertainty estimates with a deep neural network on a toy dataset.}
	\label{uncertainty}
\end{figure*}

\textbf{Quality of uncertainty.}\hspace{2mm} Finally, we turn to the issue of what effect this prior has on uncertainty and calibration. We measure calibration with the Brier score and, similar to \cite{lakshminarayanan2017simple}, the accuracy of predictions above 50\% and 90\% confidence to see whether our models tend towards overconfidence. We monitor the progression of these metrics of models with different priors through 100 epochs reported in Figure \ref{uncertainty}. As with any iteratively updating prior, we expect that it may adapt to the dataset and overfit, as is shown to be true of our stabilising prior and EB in Figure \ref{uncertainty}. As an answer to this we explore combining a regularising and stabilising prior which trains faster and results in a well calibrated model with better Brier scores than any solitary prior. 

\end{document}